# The cyclic job-shop scheduling problem

The new subclass of the job-shop problem and applying the Simulated annealing to solve it

P.V. Matrenin, V.Z. Manusov
Department of the Power Supply Systems
Novosibirsk State Technical University
Novosibirsk, Russia
pavel.matrenin@gmail.com

*Abstract*— **In the paper, the new approach to the scheduling problem are described. The approach deals with the problem of planning the cyclic production and proposes to consider such scheduling problem as the cyclic job-shop problem of the order *k*, where *k* is the number of reiterations. It was found out that planning of only one iteration of the loop is less effective than planning of the entire cycle. To the experimental research, a number of test instances of the job-shop scheduling problem by Operation Research Library were used. The Simulated Annealing was applied to solve the instances. The experiments proved that the approach proposed allows increasing the efficiency of cyclic scheduling significantly.**

I. INTRODUCTION

The combinatorial optimization is one of the most important areas of discrete mathematic because thousands of industrial tasks can be formulated as problems of the combinatorial optimization. Real-life problems often belong to NP-hard problems and have a high dimensionality. Scheduling tasks may be characterized as one of the most significant optimization problems since plans and schedules need to be arranged in all fields. Such tasks, as a rule, are usually modeled as the job-shop scheduling problem, which deals with planning multi-stage service systems. However, this view ignores the cyclic nature of the tasks. Since production processes are cycle-after-cycle often, it is necessary to make a plan not for a single execution but for multiple ones. In this paper, we deal with a modification of the job-shop scheduling problem for the cyclic production and we substantiate the claim that an optimal solution of the cyclic job-shop task is not limited to the cyclic using a solution of the usual job-shop task. For the purpose of experimental research, the Simulated Annealing (SA) algorithm is used.

The article is organized as follows. Section II describes the mathematical model of the job-shop scheduling problem and gives an example of the cyclic job-shop scheduling problem. Section III gives a brief overview of the SA algorithms. Section IV shows experimental evidence, and the conclusion describes the results.

II. STATEMENT OF THE PROBLEM

*A. Overview of the job-shop scheduling*

The job-shop scheduling problem is among the hardest combinatorial optimization problems [1, 2, 3]. In general, the job-shop problem can be described as follows: [1]. A finite set of jobs $N$ as and a finite set of machines (performers) $M$ are given as $N = \{1, 2, ..., n\}$, $M = \{1, 2, ..., m\}$. The process of servicing each job includes a number of stages (operations), so each job has the order defined through the machines during an uninterrupted time interval. The sequences of execution can be different for different jobs. There is the important rule: one machine can process at most one operation at a time. The process of system functioning can be described by specifying the schedule (a calendar plan). The schedule can be described as an allocation of the stages to time intervals on the machines. The objective of the problem is minimizing the makespan. The makespan is the total length of the schedule, i.e. it is the maximum of completion times needed for processing all stages of all jobs. Therefore, the job-shop problem is to find the shortest or quickest schedule. It is the simplified model, however, this model that can be considered as a basis for many real-life scheduling problems [1, 4].

*B. Complexity of the job-shop problem*

The job-shop problem is NP-hard generally and even among the members of this class, it belongs to the most difficult ones [2, 3]. It is known that small size instances of the problem can be solved with a reduced computational time by exact algorithms, such as brand-and-bounds, as has been shown by Lagewed, Lenstra, and Kan (1977), Carlier and Pinson (1989) [2, 5]. The job-shop scheduling problem is polynomial if it has 2 machines and no more than 2 operations per job or if the problem has 2 machines and unitary processing times [1, 2, 6]. For large instances, only heuristic algorithms achieve satisfactory results. Lenstra et al. [7] show that even some simplified versions are NP-hard. These include only 3 machines and 3 jobs; or 2 machines and no more than 3 operations per job; or 3 machines and no more than 2 operations per job; 3 machines and unitary processing times [3].

Accordingly, a brute-force enumerative algorithm for the problem has worst-case complexity $O((n!)^m)$, which is lower than the worst-case complexity for branch-and-bound algorithms [3].

There is a number of methods using priority rules for choosing the operation from a subset of yet unseduced stages [1, 9, 10]. These methods work quickly but the schedules obtained are not good enough often.

The approach of Adams, Balas and Zawack [8], named shifting bottleneck procedure has the high efficiency, but this method is laborious for applying and requires sophisticated modifications in case of changing the details of the mathematical model of the scheduling problem [2].

Now the stochastics heuristics and the meta-heuristics methods are the most effective for NP-hard optimization problem since these methods are self-organize, i.e. automatically adapting to the task solved. The Genetic algorithm, the Tabu Search, the Simulated Annealing, and the Swarm Intelligence algorithms are the most commonly used. These algorithms are used successfully for the solving the job-shop scheduling problem (the Genetic [10, 12, 13], the Tabu Search [12], the SA [2], the Swarm Intelligence [14, 15]).

*C. The mathematical model*

The mathematical model of the job-shop problem can be written as follow [12]:

- $N = \{1, 2, \ldots, n\}$ is the set of jobs.
- $M = \{1, 2, \ldots, m\}$ is the set of machines.
- $V = \{0, 1, \ldots, j+1\}$ denote the set of the operations, 0 and $j+1$ are fictions operations: start and finish.
- $A$ be the set of pairs of operations constrained by the precedence relations.
- $V_j$ be the set of operations to be performed by the machine $j$.
- $E_k \subset V_k \times V_k$ be the set of pairs of operations to be performed on the machine $k$ and which therefore have to be sequenced.
- $p_v$ and $t_v$ denote the known processing time and the unknown start time of the operation $v$.

Given this assumption, the job-shop problem can be considered as:

minimize $t_{j+1}$      $t_j - t_i = p_i$,      $(\underline{i}, j) \in A$

subject to

$t_j - t_i \geq p_i$ & $t_i - t_j \geq p_j$, $(\underline{i}, j) \in E_k$, $k \in M$       (1)

Any feasible solution to the problem (1) is called schedule

*D. The cyclic job-shop scheduling*

In this paper, we propose to analyze the job-shop optimization problems taking into account a cyclic nature of the proceeding. Such an approach implies that the setting of the classical job-shop model relates to the production of a certain set (one consignment) of products and the modified job-shop model relates to the production a number of such identical sets ($k$ consignments); $k$ is a number of sets. Let's take, for example, the job-shop task, given in [10], which describes the production of consignments with the notations $A$, $B$, $C$, $D$. The set of machines is defined as $\{R, S, T, Q\}$. The Gantt chart for this task is shown in Fig. 1.

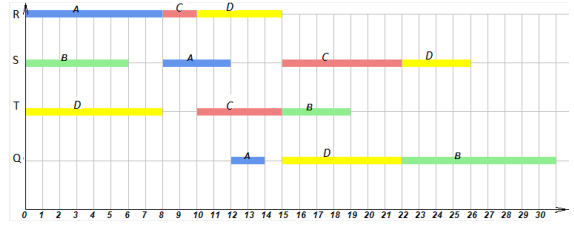

Fig. 1. The Gantt chart of a job-shop task example

The Gantt chart in Fig. 1 has the extensive vacant areas on the machine $R$ after the 16th hour and the machine $T$ after the 19th hour. If it is required to produce 2 consignments of each type, then, obviously, the execution time may be less than the double time for the planned execution shown in Fig. 1, that is less than 62 (2*31) hours. The optimal solution of this modified problem is shown in Fig. 2.

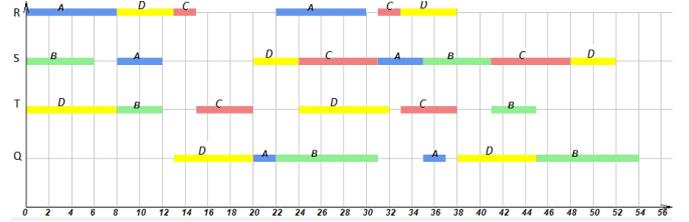

Fig. 2. The Gantt chart of a cyclic job-shop task example, $k = 2$

Fig. 2 shows that the processing time is 54 hours, in addition, the performance of the first half of the consignment goes according to plan, which differs from the plan shown in Fig. 1.

We have named the job-shop scheduling problem obtained from the classic job-shop problem by introducing requirements for the performance of all works $k$ times as the *cyclic job-shop problem of the order k*. If $k = 1$, then the *cyclic job-shop problem of the order k* is equivalent to the classic job-shop problem.

III. SOLVING THE CYCLIC JOB-SHOP PROBLEM

*A. The simulated annealing*

As it is shown in Fig. 1 and Fig, 2, the solution to the job-shop task of the order $k$ can be more effective than simply copying the solution result of the job-shop task $k$ times. In order to conduct research on the various tasks, it is necessary to choose an efficient algorithm for obtaining solutions that are close to optimum solutions. Since the dimension of the job-shop task of the order $k$ is much higher than the classical task, and the deviation of the solution from optimum level by several percentages is not crucial for scheduling purposes, it is advisable to apply the SA algorithm. This algorithm enables us to quickly obtain optimum solutions; it is especially effective in combinatorial problems [2, 16].

The inspiration for the SA comes from the physical process of cooling molten materials down to the solid state [16, 17]. The energy state of a system is described by the energy state of each particle of the system. A particle's energy state changes randomly and a probability of moving depends on the temperature of the system. The probability of moving from a high-energy state to a lower-energy state is very high. In the oppositional case, the probability is less but is nonzero. The higher the temperature, the more likely energy moving will occur.

Any combinatorial optimization problem can be considered as minimizing the total energy, i.e. as a task of a search minimum-energy state. The random transitions (changes of a current solution) generated according to the given probability distribution mimics the physics cooling process to solve a combinatorial optimization problem. At first, the SA uses randomness to explore the search space of all possible solutions widely, so the probability of accepting a negative moving should be high. The cooling process regulated by the following parameters [16]:

- Initial system temperature, $t_1$.

- Temperature decrement function, typically $t_{i+1} = \alpha \cdot t_i$, where $0.0 < \alpha < 1$.

- A number of iterations between temperature change (*step_temp*).

- Acceptance criteria. A criterion is to accept any transition from solution $s_{current}$ to solution $s_{new}$ when $s_{new}$ better than $s_{current}$, and also accept a negative transition whenever $\exp(-(criterion(s_{current}) - criterion(s_{new})) / kt \cdot t_i) \geq r$, where r is a random number ($0 \leq r < 1$).

- Stop criteria. After evaluating a certain number of iterations, the search is terminated.

A number of studies have proved the high efficiency of the SA, such as [2, 16, 18, 19].

*B. The realization of the SA algorithm*

The realization of the SA algorithm for solving job-shop task showed from below using the pseudocode.

*Start annealing algorithm*
{
  /* initialization */
  temperature ← INITIAL_TEMPERATURE
  solution ← initialize()
  current_value = schedule_length(solution)
  counter_steps ← 0

  *while* (counter_steps < COOLING_STEPS)
  {
    temperature ← temperature* COOLING_FRACTION
    start_value = current_value
    counter_steps_temp ← 0

    *while*(counter_steps_temp < STEPS_TEMP)
    {
      /* pick randomly two elements of a schedule to swap
      r1 ← random_integer(1, N)
      r2 ← random_integer(1, N)

      /*create a new schedule and find it's length */
      solution ← swap_schedule(solution, r1, r2)
      new_value = schedule_length(solution)
      delta = new_value − current_value

      *if* (delta < 0) /*find a better solution*/
      {
        current_value = new_value
      }
      *else* /*find a worse solution, use a randomize chose */
      {
        ex = exp((-delta/current_value)/(KT*temperature))
        *if* (ex > randon_float(0,1)) /*accept new solution */
        {
          current_value = new_value
        }
        *else* /* reject */
        {
          solution ← swap_schedule(solution, r2, r1)
        }
      }
      counter_steps_temp ← counter_steps_temp + 1
    }

    /* restore temperature if progress has been mad */
    *if* ((current_value - start_value) < 0.0)
    {
      temperature ← temperature/COOLING_FRACTION;
    }

    counter_steps ← counter_steps + 1
  }
}
End annealing algorithm

Table 1 shows the values of coefficients used in this research.

TABLE I. THE VALUES OF THE SA COEFFICIENTS

| Parameter | Value |
|---|---|
| INITIAL_TEMPERATURE | 1.0 |
| COOLING_STEPS | 3000 |
| COOLING_FRACTION | 0.97 |
| STEPS_TEMP | 3000 |
| KT | 0.01 |

## IV. SOLVING THE CYCLIC JOB-SHOP PROBLEM

### A. Test instances used

A number of test instances were used as a benchmark for the experiments. The test tasks were selected from OR-Library [20] ("abz" – Adams, Balas, Zawack, "ft" – Fisher and Tomson, la - Lawrence), the real industrial task from [10] ("sk" – Sekaev), and example showed in Figure 1. The complexities of the tasks are listed in Table 2.

TABLE II. THE COMPLEXITIES OF THE TEST INSTANCES

| Task | n, number of jobs | m, number of machines | $l_{max}$, max number of operations from all jobs |
|---|---|---|---|
| abz6 | 10 | 10 | 10 |
| ft06 | 6 | 6 | 6 |
| ft10 | 10 | 10 | 10 |
| ft20 | 20 | 5 | 5 |
| la01 | 10 | 5 | 5 |
| la02 | 10 | 5 | 5 |
| la03 | 10 | 5 | 5 |
| la04 | 10 | 5 | 5 |
| la05 | 10 | 5 | 5 |
| la06 | 15 | 5 | 5 |
| la07 | 15 | 5 | 5 |
| la08 | 15 | 5 | 5 |
| la09 | 15 | 5 | 5 |
| la10 | 15 | 5 | 5 |
| la11 | 20 | 5 | 5 |
| la12 | 20 | 5 | 5 |
| la13 | 20 | 5 | 5 |
| la14 | 20 | 5 | 5 |
| la15 | 20 | 5 | 5 |
| la16 | 10 | 10 | 10 |
| la17 | 10 | 10 | 10 |
| la18 | 10 | 10 | 10 |
| la19 | 10 | 10 | 10 |
| la20 | 10 | 10 | 10 |
| la21 | 15 | 10 | 10 |
| Fig.1 | 4 | 4 | 4 |
| sk | 3 | 6 | 23 |

### B. The results comparison

For each test instance, a quasi-optimal solution was carried out by the SA in terms of the classical formulation (1st order), and then as well as modifications, i.e. cyclic job-shop problem 2nd and 4th orders. To compare our approach with the reiterations of the schedule obtained for 1st order task, we took the best-known solutions from [11] and multiplied them the number of reiterations (1, 2 and 4). The schedule's lengths, obtained by other methods are shown in Table 3. To be specific, let's assume that the schedule's lengths are measured in hours.

In Table 3, the following notations are used:

- Best 1 – quasi-optimal result from [11], which can be considered as the best possible solution;
- SA 1 – the result obtained by the SA algorithm;
- Best 2 – the best result of solving the task, multiplied by 2, i.e. obtained by a simple repetition of the plan for the task of the first order;
- 

TABLE III. COMPARISON THE REITERATIONS AND SOLVING OF THE CYCLIC JOB-SHOP PROBLEM

| Task | Best 1 | SA 1 | Best 2 | SA 2 | Best 4 | SA 4 | Dif. % |
|---|---|---|---|---|---|---|---|
| abz6 | 943 | 943 | 1886 | 1810 | 3772 | 3482 | **7.69** |
| ft06 | 55 | 55 | 110 | 103 | 220 | 197 | **10.45** |
| ft10 | 930 | 937 | 1860 | 1661 | 3720 | 3112 | **16.34** |
| ft20 | 1165 | 1178 | 2330 | 2280 | 4660 | 4484 | **3.78** |
| la01 | 666 | 666 | 1332 | 1332 | 2664 | 2664 | **0** |
| la02 | 655 | 655 | 1310 | 1290 | 2620 | 2560 | **2.29** |
| la03 | 597 | 597 | 1194 | 1176 | 2388 | 2352 | **1.51** |
| la04 | 590 | 590 | 1180 | 1115 | 2360 | 2186 | **7.37** |
| la05 | 593 | 593 | 1186 | 1186 | 2372 | 2372 | **0** |
| la06 | 926 | 926 | 1852 | 1852 | 3704 | 3704 | **0** |
| la07 | 890 | 890 | 1780 | 1759 | 3560 | 3497 | **1.77** |
| la08 | 863 | 863 | 1726 | 1726 | 3452 | 3452 | **0** |
| la09 | 951 | 951 | 1902 | 1902 | 3804 | 3804 | **0** |
| la10 | 958 | 958 | 1916 | 1916 | 3832 | 3832 | **0** |
| la11 | 1222 | 1222 | 2444 | 2444 | 4888 | 4888 | **0** |
| la12 | 1039 | 1039 | 2078 | 2078 | 4156 | 4156 | **0** |
| la13 | 1150 | 1150 | 2300 | 2300 | 4600 | 4600 | **0** |
| la14 | 1292 | 1292 | 2584 | 2584 | 5168 | 5168 | **0** |
| la15 | 1207 | 1207 | 2414 | 2414 | 4828 | 4828 | **0** |
| la16 | 945 | 946 | 1890 | 1712 | 3780 | 3272 | **13.4** |
| la17 | 784 | 784 | 1568 | 1501 | 3136 | 2946 | **6.06** |
| la18 | 848 | 848 | 1696 | 1621 | 3392 | 3156 | **6.96** |
| la19 | 842 | 848 | 1684 | 1639 | 3368 | 3138 | **6.83** |
| la20 | 902 | 907 | 1804 | 1722 | 3608 | 3338 | **7.48** |
| la21 | 1046 | 1074 | 2092 | 2043 | 4184 | 4013 | **4.27** |
| Fig.1 | 31 | 31 | 62 | 54 | 124 | 102 | **17.8** |
| sk | 657.55 | 657.55 | 1315.1 | 1284.05 | 2630.2 | 2539.4 | **3.45** |

- SA 2 – solution to the task of the 2nd order obtained by the SA algorithm;
- Best 4 – the best result of solving the task, multiplied by 4, i.e. obtained by a simple repetition of the plan for the task of the first order;
- SA 4 – solution to the task of the 4th order obtained by the SA algorithm;
- Dif – difference between "Best 4" and "SA 4", this value shows as far the consideration job-shop problem as a cyclic problem more efficiently than a simple repetition of the solution of the first-order job-shop problem.

The maximum difference between results "Best 4" and "SA 4" is 17.8%, the average difference is 4.35%. However, is we ignore the tasks, which have less than ten machines (la01-15, ft20, Fig.1), and then the average difference will be 8.3% or 97 hours. It should be noted high-speed operation of the SA algorithm. Thus, to solve the tasks under consideration the algorithm required from a few seconds to 10 minutes (the SA algorithm were implemented on a 2.4 GHz Intel CPU i7 using C++ language).

It is obvious that increasing the number of repetitions of a production cycle leads to increasing the difference of effectiveness. For example, Table 4 shows schedules' lengths for the task la20. However, for orders 6-10 we had to increase the number of the SA steps to 6000 (COOLING_STEPS and STEPS_TEMP) because the complexity of the task proved to be too high.

TABLE IV. SCHEDULES' LENGTHS FOR LA20

| Order | Reiteration the first order task | Solving as the cyclic task | Difference, hours | Difference, % |
|---|---|---|---|---|
| 1 | 902 | 907 | -3 | -0.55% |
| 2 | 1804 | 1722 | 82 | 4.54% |
| 4 | 3608 | 3338 | 270 | 7.48% |
| 6 | 5412 | 4895 | 517 | 9.55% |
| 8 | 7216 | 6497 | 719 | 9.96% |
| 9 | 8118 | 7401 | 717 | 8.80% |
| 10 | 9020 | 8113 | 907 | 10.1% |

## Conclusion

Since scheduling often has a cyclic nature, in some situations it is necessary to consider drawing up plans in terms of their reiterations. Planning a single iteration with subsequent reiterations may be inefficient in comparison with a plan prepared for all the iterations of the cycle as it is proposed in this paper. The experiments show that in the case of four and more reiterations the second approach is significantly more effective. Therefore, we propose to consider such scheduling problem as the *cyclic job-shop problem of the order k*, where $k$ is the number of iterations. In the experiments, the average difference of the effectiveness proved to be about 8% for four iterations ($k = 4$). Increasing the number of reiterations results in the growth of the difference. It first increases linearly, then goes into saturation, since the computational complexity of the task increases exponentially.

The proposed approach increases efficiency and it depends on the conditions of a specific problem. The increasing computational time for preparing a schedule is usually insignificant in relation to the time saving due to a more efficient schedule. The results of this paper may be practically applied to solve scheduling tasks in the field of the multi-stage service systems.

The Simulated Annealing algorithm can be easy implemented for different optimization problem and it demonstrates a high-performance for the scheduling problems.